\documentclass[
]{ceurart}

\usepackage{hyperref}
\makeatletter
\providecommand{\doi}[1]{
  \begingroup
    \let\bibinfo\@secondoftwo
    \urlstyle{rm}
    \href{http://dx.doi.org/#1}{
      doi:\discretionary{}{}{}
      \nolinkurl{#1}
    }
  \endgroup
}
\makeatother

\usepackage{caption}
\usepackage{comment}

\usepackage{subcaption}
\usepackage{graphicx}
\usepackage{grffile}  

\usepackage{wrapfig}

\usepackage{svg}
\usepackage{capt-of}
\usepackage{booktabs}
\usepackage{varwidth}
\usepackage{float}

\usepackage{todonotes}

\begin{document}

\copyrightyear{2021}
\copyrightclause{Copyright for this paper by its authors.
  Use permitted under Creative Commons License Attribution 4.0
  International (CC BY 4.0).}

\conference{3rd International Workshop on Data meets Applied Ontologies, September 2021 -  https://daoxai.inf.unibz.it}

\title{Automated and Explainable Ontology Extension Based on Deep Learning: A Case Study in the Chemical Domain}

\address[1]{
  Otto von Guericke University Magdeburg,          
   Universitaetsplatz 2,
   39106 Magdeburg,
   Germany 
}

\address[2]{
  Free University of Bozen-Bolzano,          
  piazza Universit\`{a}, 1
  39100 Bolzano,                               
  Italy                                  
}

\address[3]{
  University College London,
  Gower Street,
  London WC1E 6BT,
  United Kingdom
}

\author[1]{Adel Memariani}[
]

\author[1]{Martin Glauer}[
]

\author[1,2]{Fabian Neuhaus}[
]

\author[1]{Till Mossakowski}[
]

\author[1,3]{Janna Hastings}[
]

\begin{abstract}
Reference ontologies provide a shared vocabulary and knowledge resource for their domain. Manual construction enables them to maintain a high quality, allowing them to be widely accepted across their community. However, the manual development process does not scale for large domains. 
We present a new methodology for automatic \textit{ontology extension} and apply it to the ChEBI ontology, a prominent reference ontology for life sciences chemistry. We trained a Transformer-based deep learning model on the leaf node structures from the ChEBI ontology and the classes to which they belong. The model is then capable of automatically classifying previously unseen chemical structures. The proposed model achieved an overall F1 score of 0.80, an improvement of 6 percentage points over our previous results on the same dataset. Additionally, we demonstrate how visualizing the model's attention weights can help to explain the results by providing insight into how the model made its decisions.

\end{abstract}

\begin{keywords}
  ontology extension \sep 
  ontology generation \sep
  ontology learning \sep 
  chemical ontology \sep
  Transformers \sep
  automated classification \sep
  transfer learning \sep
  multi-label classification
\end{keywords}

\maketitle

\setlength{\tabcolsep}{4pt}

\setcounter{page}{2}
\pagenumbering{arabic}

\section{Introduction}
\label{sec:intro}
Ontologies represent knowledge in a way that is both accessible to humans and is machine interpretable. Reference ontologies provide a shared vocabulary for a community, and are successfully being used in a range of different domains. Examples include the OBO ontologies in the life sciences \cite{smith2007obo}, the Financial Industry Business Ontology for the financial domain \cite{allemang2021}, and the Open Energy Ontology in the energy domain \cite{booshehri2021introducing}. While these ontologies differ in many respects, 
they share one important feature: they are manually created by experts using a process by which each term is manually added to the ontology including a textual definition, relevant axioms, and ideally some additional documentation. 
Often, this process involves extensive discussions about individual terms. Hence, developing such ontologies is a time-intensive and expensive process. This leads to a challenge for ontologies that cover a large domain. 

For example, the  
ChEBI (Chemical Entities of Biological Interest) ontology \cite{hastings_chebi_2016} is  the largest and most widely used ontology for the domain of biologically relevant chemistry in the public domain. 
It currently (as of June 2021) contains  59,122  fully  curated classes, which makes it large in comparison to other reference ontologies.
ChEBI is largely manually maintained by a team of expert curators. This is an essential prerequisite for its success, because it enables it to capture the terminology and classification logic shared by chemistry experts.
However, the number of chemicals covered by ChEBI is dwarfed by the 110 million chemicals in the PubChem database \cite{wang_pubchem_2009}, which itself is not comprehensive. 
The manually curated portion of ChEBI only grows at a rate of around 100 entries per month, thus will only ever be able to cover a small fraction of the chemicals that are in its domain.

ChEBI tries to navigate this dilemma by extending the manually curated core part of the ontology automatically 
using the ClassyFire tool \cite{djoumbou_feunang_classyfire_2016}.
This approach has tripled ChEBI's coverage to 165,000 classes (as of June 2021). However, there are limitations to this approach. Firstly, 
ClassyFire  uses a different underlying classification approach to ChEBI (e.g. conjugate bases and acids are not distinguished), thus, mapping to ChEBI loses classification precision.
More importantly, ClassyFire is rule-based and while the extension of the ontology is automated, the creation and curation of the ClassyFire's rules is not. This limits the scalability of this approach. 

Somewhat inspired by ChEBI's workflow, we suggest navigating the ontology scaling dilemma by using a new kind of approach to  \emph{ontology extension}, which  transfers the design decisions of an existing ontology analogously to new classes and relations. Our starting point is an existing, manually curated reference ontology. We suggest the use of machine learning methods to learn some of the criteria that 
the ontology developers adopted in the development of the ontology, and then use the learned model to extend the ontology to 
entities that have not been covered by the manual ontology development process yet. We will illustrate this approach in this paper for the chemistry use case by training an artificial neural network (with a Transformer-based architecture) to automate the extension of ChEBI with new classes of chemical entities. The approach has several benefits: since it builds on top of the existing ontology, the extension will preserve the manually created consensus. Moreover, the model is trained solely on the content of the ontology itself and does not rely on any external sources. 
Finally, as we will see, the chosen architecture allows explanation of the choices of the neural network, and, thus to validate the trained model to some degree by manual inspection. 

In the next two sections we discuss related work and the overall methodology that we are using to train a model for classifying new classes of chemical entity as subclasses of existing classes in ChEBI.  

\section{Related Work}
In this paper, we present a methodology for ontology extension, which can be considered as a kind of ontology learning. Ontology learning has been an active area of research for more than two decades \cite{assadi:hal-01617868,maedche2001ontology,biemann2005ontology,asim2018,ozaki2020learning} and
 a number of automated ontology generators have been developed.
A recent publication \cite{ozaki2020learning} defined a list of six \textit{desirable goals} for ontology learning methods: they should support expressive languages, require small amount of time and training data, require limited or no human intervention, support unsupervised learning, handle inconsistencies and noise, and their results should be interpretable.
Note that in this paper, we use the terms \textit{explainability} and \textit{interpretability} interchangeably. 
Sometimes explainability is considered to be a stronger form of interpretability \cite{gilpin2018explaining}. 

The fundamental make-up of the resulting ontologies varies widely -- in part due to different notions of what constitutes an ontology.
A survey-based study by Biemann \cite{biemann2005ontology} defines three classes of ontologies: \textit{formal}, \textit{prototype-based} and \textit{terminological} ontologies.
Most early and data-driven approaches resulted in prototype-based ontologies, in which concepts are not defined in natural language or by logical formulae, but solely by their members.
New concepts are often derived from metric-based aggregations such as hierarchical clustering \cite{karoui2007contextual}.
The quality of the resulting classification depends strongly on the chosen representation of individuals and the criteria for similarity, and may not agree with distinctions that are used by domain experts.

Advances in natural language processing led to a different class of approaches - the \textit{terminological} ontologies.
Here, artificial intelligence is used to analyse corpora of relevant literature in order to extract important terms and their relations.
Yet, these approaches reflect rather than resolve the inherent ambiguities and differences in language use that exist within different communities of domain experts or even within single communities.
The resolution of these ambiguities is an essential part of the ontology development process that involves extensive in-depth communication with and between domain experts \cite{booshehri2021introducing}.
Finally, formal  ontologies place a strong emphasis on definitions distinguishing entities, and a rich logical axiomatisation that yields a powerful foundation for reasoning and data integration \cite{xiao2018ontology, dessimoz_primer_2017, fikes2004owl}.

While the majority of existing approaches in ontology learning focus on creating new ontologies from scratch, the ones that are dedicated to ontology extension use the ontology as a seed to identify terms that are important for the target domain \cite{liu2005semi, althubaiti2020combining, zhou2016research, barchi2014never, schutz2005relext, petrova2015formalizing}. These are  used to guide approaches that are similar to those that are applied to learn ontologies 'from scratch'. Hence, the resulting extensions are not necessarily based on the principles that have been employed to develop the ontology in the first place, and may potentially introduce biases from the literature into the ontology. Some approaches involve several manual steps, in which experts evaluate concepts and related phrases to sort out these potential issues \cite{li2019method}. Involving human experts has the advantage of providing quality control, but is labour-intensive and costly.

Our approach differs from the existing work in that it employs machine learning techniques but does not rely on text corpora. Rather, it relies only on the content of the ontology that is being extended, in particular on structured annotations.   

Our specific application domain is chemical ontology. One characteristic of chemical ontologies is the fact that many classes of chemical entities are annotated with information about their chemical structure. 
Particularly important for our purposes are annotations in  the  
Simplified Molecular-Input Line-Entry System (SMILES) \cite{weininger1988smiles}, which is used to represent chemical entities as a linear sequence of characters. 
The SMILES notation is analogous to a language to describe atoms and their bonds within a chemical entity.

In our approach, we train a deep learning classifier that is based on the ChemBERTa \cite{chithrananda2020chemberta} architecture. This chemistry-focused Transformer-based architecture has been successfully employed for toxicity prediction. In the context of this work, it has been trained for the first time on the structural annotations of an existing chemical ontology. The learning method biases the classifier towards the ontology's internal structure, yielding a model that is in line with the domain experts' conceptualisation as represented in the existing ontology.
The resulting model is then used to integrate previously unseen classes into the ontology.

This is a novel approach to the problem of chemical classification, which task has historically been approached in multiple different ways \cite{hastings_structure-based_2012}. 
Solutions that involve deep-learning methods were successfully employed for many other applications in chemistry \cite{mater2019deep}, such as the prediction of properties of chemicals \cite{goh2017smiles2vec} or reaction behaviour \cite{coley2019graph}.
Yet, the automated classification of chemicals using deep learning according to an existing ontology has been largely unexplored. 
The ClassyFire tool \cite{djoumbou_feunang_classyfire_2016} is at the time of writing the most comprehensive method for structure-based automated chemical ontology extension.
However, it uses a rule-based and algorithmic implementation that is cumbersome to maintain and is not able to adapt as the underlying ontology changes.

In our previous work \cite{hastings2021learning}, we have evaluated several classifiers for this task, including a long short-term memory (LSTM) model which was the best-performing overall.
The results of this effort were satisfactory as a whole, but several specific limitations were identified.
In particular, the model failed to provide any prediction for a subset of input molecules, and the system as a whole offered no explainability.
The current contribution harnesses a Transformer-based architecture and describes how the attention weights of the resulting model can provide insights into how the model made its decisions.
Furthermore, by using transfer learning, a broader applicability of this data- and compute-hungry method becomes computationally more feasible.

\section{Methodology}
\label{sec:methodology}

Our goal is to train a system that automatically extends the ChEBI ontology with new classes of chemical entities (such as molecules) based on the design decisions that are implicitly reflected in the structure of ChEBI. 
Thus, for our work we take the `upper level' of the ontology, which contains generic distinctions, as given. Moreover, ChEBI is relatively weakly axiomatised, consisting largely of a taxonomy supplemented by existentially restricted relationships. All chemical entities in ChEBI are represented as \textit{classes} and there are no individuals in the ontology. Thus, the ontology extension task consists of adding classes and subsumption relationships. 

Our focus is the extension of the ChEBI ontology with classes of chemical entities that may be characterised by a SMILES string, i.e., they are associated with a specific chemical structure. Chemical structures can typically only be specified for relatively specific types of chemical entity, thus, although these classes are not necessarily leaf nodes in the ontological hierarchy, they nevertheless tend to be in the `lower' (more specific) part of the hierarchy. 

The learning task for ontology extension may, thus, be characterised as follows: \textit{ Given a class of chemical entities (characterised by a SMILES string), what are its optimal direct superclasses in ChEBI}? 

While our goal is -- from an ontological point of view -- to extend the ChEBI ontology with new classes (i.e., adding new subsumptions), from a machine learning perspective we turned this problem into a classification task, for which we prepare an appropriate learning dataset from the ontology. 

Hierarchical chemical classifications should group chemical compounds in a scientifically valid and meaningful way \cite{hastings_structure-based_2012,bobach2012automated}. Each chemical entity has many structural features which contribute to its potential structure-based classification and structures that determine different classes may occur in a single molecule. Thus, ChEBI contains classes that overlap (i.e. share members). The ChEBI ontology provides two separate classification hierarchies for the chemical entities: one based on their structures and another based on their functions or uses. In the current work, we focus on the structure-based sub-ontology. Entities in the structure-based sub-ontology are often associated with specifications of their molecular structures, particularly -- but not exclusively -- the leaf nodes within the classification hierarchy. In ChEBI, a chemical entity with a defined structure can be the classification parent for another structurally defined entity, since all entities are classes according to the ChEBI ontology, and there can be different levels of specificity even amongst structurally defined classes. To formulate a \textit{supervised} machine learning problem, however, we need to create a distinction between those entities with chemical structures that form the input for learning, and the chemical classes that they belong to that form the learning target. This distinction is created by sampling structurally defined entities only from the ontology leaf nodes. 

As mentioned above, the SMILES notation is analogous to a language to describe atoms and their bonds within a chemical structure. Intuitively, this leads to a correspondence between the processing of chemical structures in this type of representation, and natural language processing \cite{schwaller2018found}. Therefore, architectures that have been successfully applied to language-based problems can also be employed for this multi-label prediction task. One of these successful architectures is \textit{Bidirectional Encoder Representations from Transformers} (BERT) \cite{devlin2018bert} -- a precursor of the \textit{Robustly optimized BERT}, (RoBERTa)  \cite{liu2019roberta} architecture that our approach is based on. The BERT architecture offers a learning paradigm that enables pre-training the model on unlabeled data and then fine-tuning it for the ultimately desired task. Fine-tuning can be done by adding one additional layer to the pre-trained model, without requiring major modifications to the model's architecture. BERT is pre-trained on two unsupervised tasks: Masked Language Modeling (MLM), in which some tokens are randomly removed from the input sequences and the model will train to predict the masked tokens, and Next Sentence Prediction (NSP), a binary classification task that predicts whether or not the second sentence in the input sequence follows the first sentence in the original text. The RoBERTa model is an extension of the BERT model and it offers several improvements with minor changes in the pre-training strategy. The RoBERTa model does not include the NSP part of BERT, and it employs a dynamic masking approach as a replacement of the original masking scheme of BERT. While the original BERT model only applies masking once during data preprocessing, the RoBERTa model dynamically changes the masking pattern on each training sequence in every epoch. As a result, the model gets exposed to different versions of the same input data with masks on various locations.

\begin{figure}
  \centering
  \includegraphics[scale=0.47]{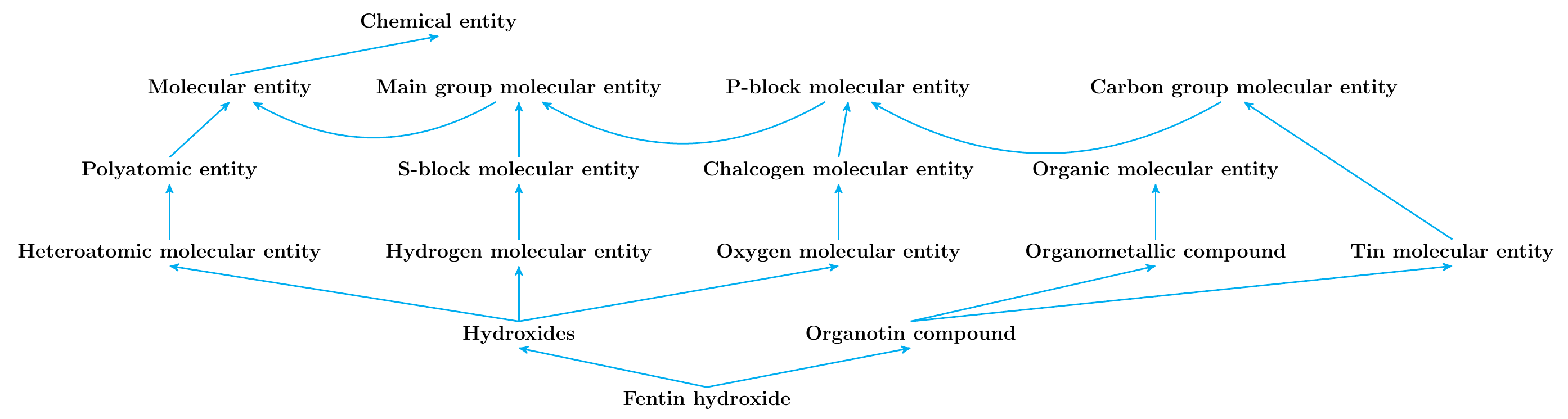}
  \caption{\textit{Fentin hydroxide} and its hierarchical classes. Blue lines indicate the \textit{sub-class} relationships.}
  \label{Fig:Fentin_hydroxide}
\end{figure}

Since chemical structures in ChEBI typically belong to several ontology classes, the problem of automated chemical entity categorization can be viewed as a \textit{multi-label} prediction task. Figure~\ref{Fig:Fentin_hydroxide} shows the \textit{fentin hydroxide} molecule and its parents in the ChEBI ontology: \textit{organotin compound} and \textit{hydroxides}. 

Our approach\footnote{\url{https://github.com/adelmemariani/chebi-roberta}} pre-trains a RoBERTa model on SMILES strings, and then predicts multiple chemical
class memberships. The overall architecture is shown in Figure~\ref{fig:ontolearn-arch}. This architecture is similar to the one that was used for molecular property prediction in \citet{chithrananda2020chemberta}.

\subsection{Dataset}
\label{sec:dataset} 
\begin{figure}
  \centering
  \begin{minipage}[b]{0.48\textwidth}
    \includegraphics[width=\textwidth]{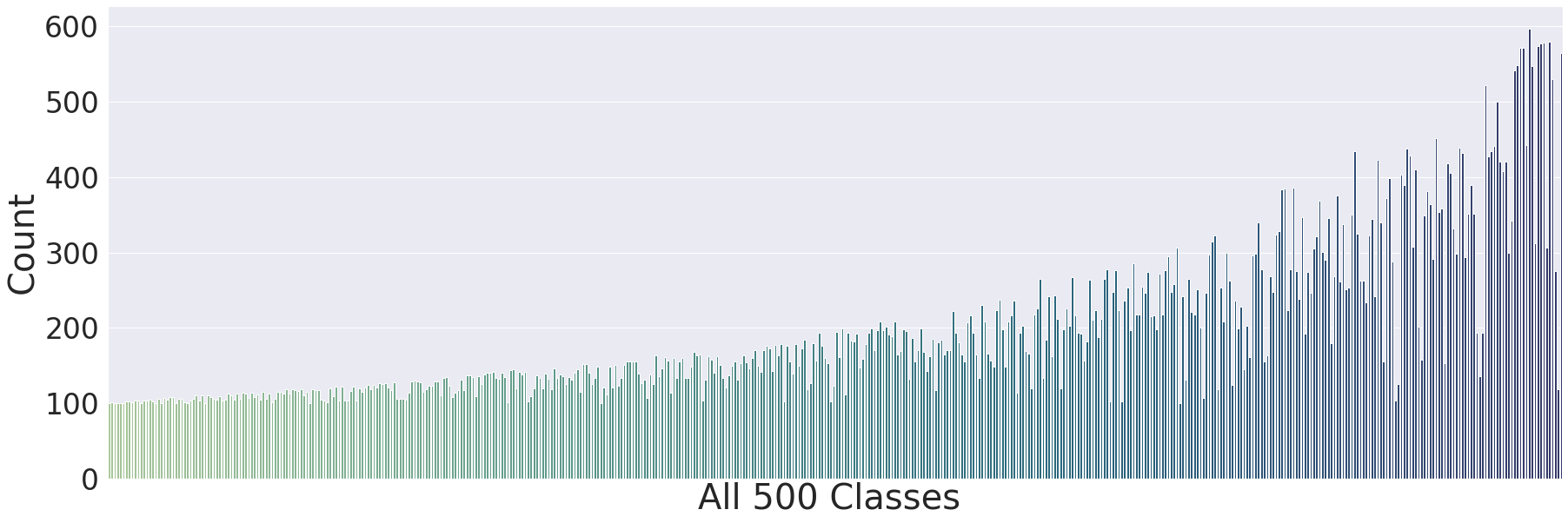}
  \end{minipage}
  \hfill
  \begin{minipage}[b]{0.48\textwidth}
    \includegraphics[width=\textwidth]{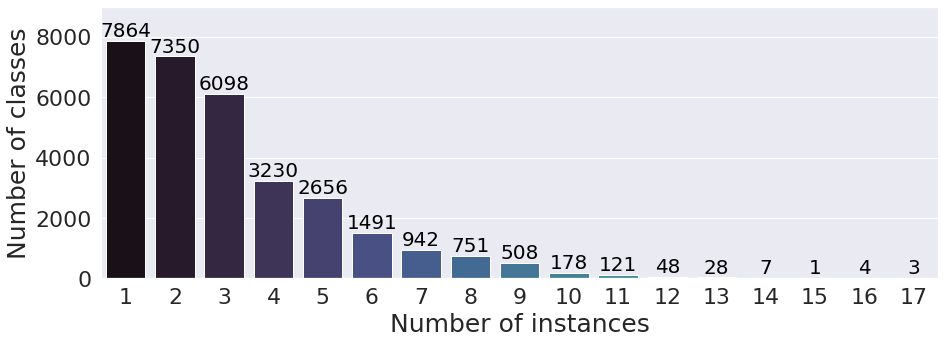}
  \end{minipage}
  \caption{Left: class counts in the dataset. Right: Number of members per number of assigned classes}
  \label{Fig:dataset}
\end{figure}

To use the existing ontology classification as input to the learning task, the ontology first has to be transformed into an appropriate form. The ontology classification is inherently unbalanced, as different classes have different numbers of members and are partially overlapping. It is therefore necessary to define a \textit{sampling strategy} to select leaf node entities and classes to minimize the impact on the training. In order to be able to compare our results to our earlier findings, we have used the same dataset\footnote{\url{https://doi.org/10.5281/zenodo.4519815}} and sampling strategy as was used in \citet{hastings2021learning}. Using only the hierarchical sub-class relations in the ChEBI ontology, this dataset was created by randomly sampling leaf node molecular entities from higher-level classes that they are subclasses of, using an algorithm that aimed to minimize (as far as possible) class overlap, described in Section 3 of \cite{hastings2021learning}. The resulting dataset contained a total of 500 molecule classes and 31,280 molecules. Despite these balancing measures, it still suffers from certain imbalances. Figure~\ref{Fig:dataset} (left) illustrates the number of times each class has appeared in the training and test datasets. As illustrated, some of the classes appeared more frequently than others. Figure~\ref{Fig:dataset} (right) shows the number of members per number of associated classes. For example, 7,864 members have just one assigned class, whereas three members have 17 classes assigned. To train, validate and test our model, we divided the dataset into three subsets; a training set containing 21,896 molecules, a validation set of 2,815 molecules, and a test set of 6,569 molecules.

\subsection{Input Encodings}
Tokenization is a pre-processing step used to create a vocabulary from textual data. It is applicable at the character, word, or sub-word level. Pre-trained large-scale word embeddings such as Word2Vec \cite{mikolov2013efficient} and GloVe \cite{pennington2014glove} employ word tokenization to generate vector representations for words that can encapsulate their meanings, semantic connections, and the contexts in which they are used. Transformer-based models rely on a subword tokenization algorithm that counts the occurrences of each character pair in the dataset and incrementally adds the most frequently occurring pairs to the vocabulary. In our previous work, \citet{hastings2021learning}, we used two strategies to encode the input sequences for the LSTM model: a character-level tokenization and an atom-wise tokenization, where letter combinations that represent an atom were encoded as a token. In the current work, we use the Byte Pair Encoding (BPE) algorithm as a sub-word tokenization method with a RoBERTa architecture.

\begin{table}
  \tiny
  \centering
  \begin{varwidth}[b]{0.4\linewidth}
    \centering
    \begin{tabular}{lllll}
    \toprule
    \textbf{Parameter}  & \textbf{Value}\\
    \midrule
    Number of attention heads & 12  \\
    Number of hidden layers & 6  \\
    Dropout for attention probabilities & 0.1  \\
    Activation function in the encoder & gelu  \\
    Activation for the classification layer & sigmoid  \\
    Number of epochs in pre-training & 100\\
    Number of epochs in fine-tuning & 30\\
    Masked language modeling probability\!\!\!\! & \%15\\ 
    Batch size & 4 \\
    Loss function for pre-training & BCELoss \\
    Loss function for fine-tuning & BCEWithLogitsLoss \\
    Optimizer & Adam with weight decay \\
    Number of vocabularies (tokens) & 1395  \\
    Number of trainable parameters & 45,577,728  \\ 
    Tokenizer & BPE  \\
    \bottomrule
    \end{tabular}
    \caption{(Hyper-)Parameters of the model}
    \label{tbl:hp}
  \end{varwidth}
  \hfill
  \begin{minipage}[b]{0.55\linewidth}
    \centering
    \includegraphics[scale=0.45]{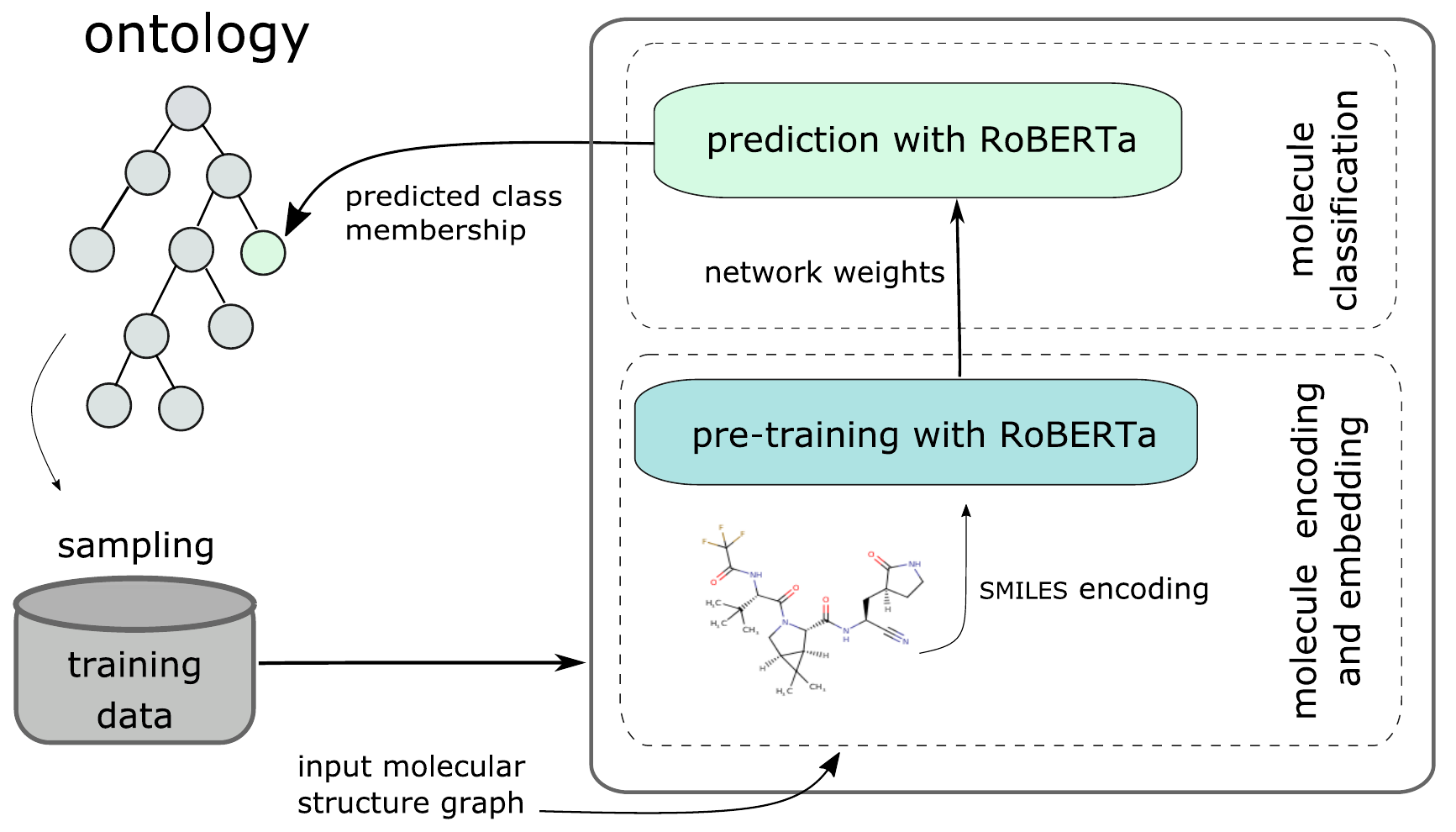}
    \captionof{figure}{Architecture of our ontology extension approach}
    \label{fig:ontolearn-arch}
  \end{minipage}
\end{table}

\subsection{Experiment}
\begin{figure}
  \centering
  \begin{minipage}[b]{0.3\textwidth}
    \centering
    \includegraphics[width=\textwidth]{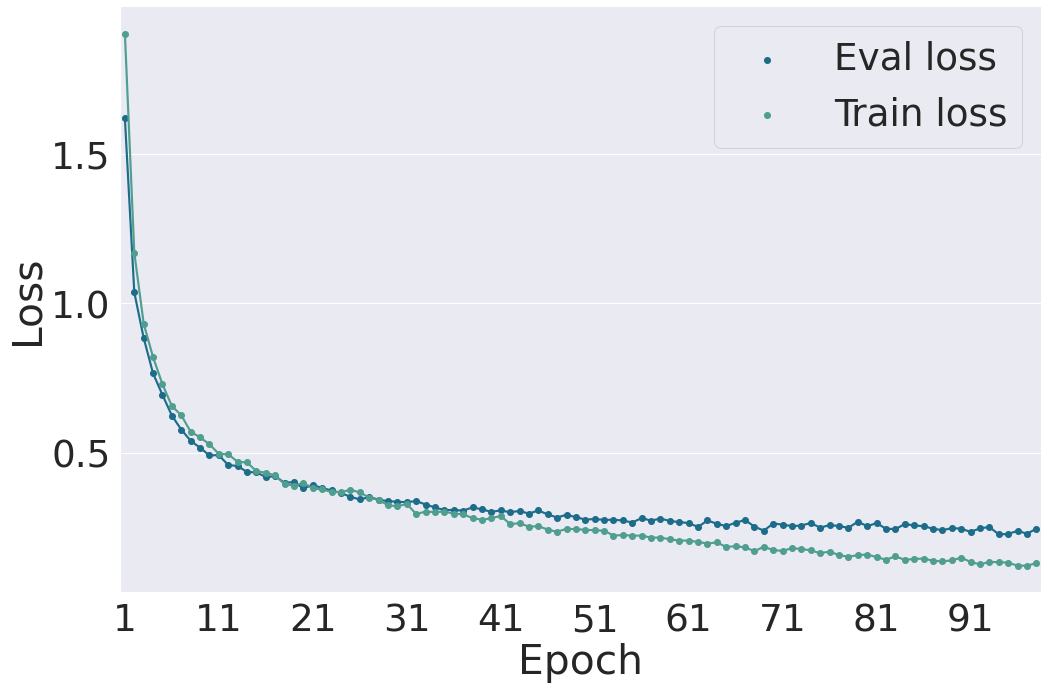} \\
    (a)
  \end{minipage}
  \hfill
  \begin{minipage}[b]{0.3\textwidth}
    \centering
    \includegraphics[width=\textwidth]{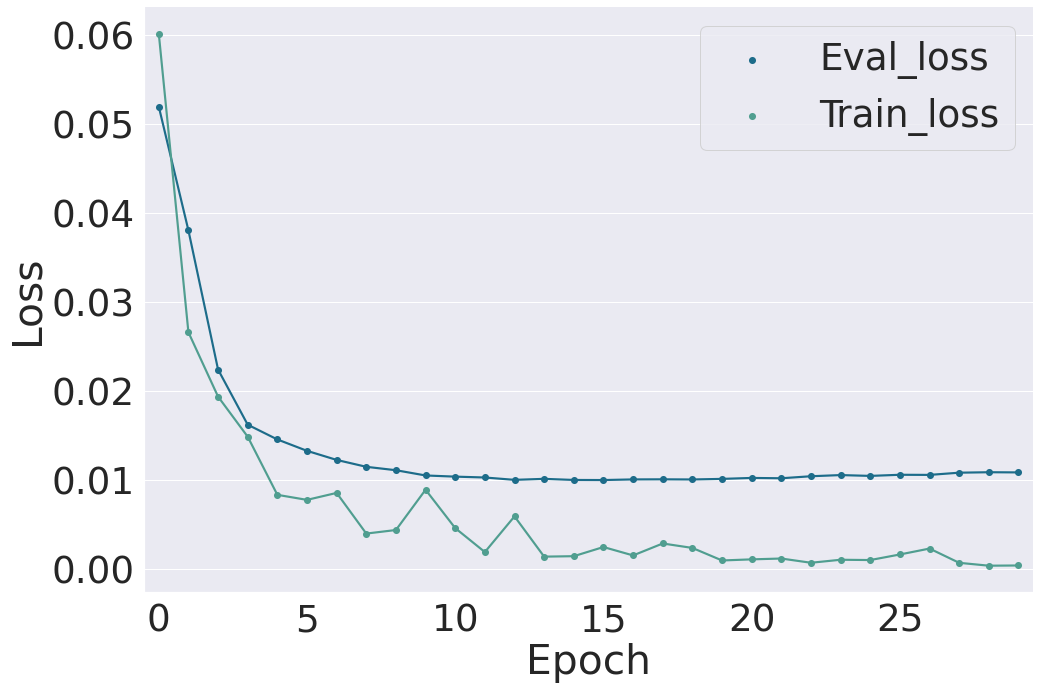} \\
    (b)
  \end{minipage}
  \hfill
  \begin{minipage}[b]{0.3\textwidth}
    \centering
    \includegraphics[width=\textwidth]{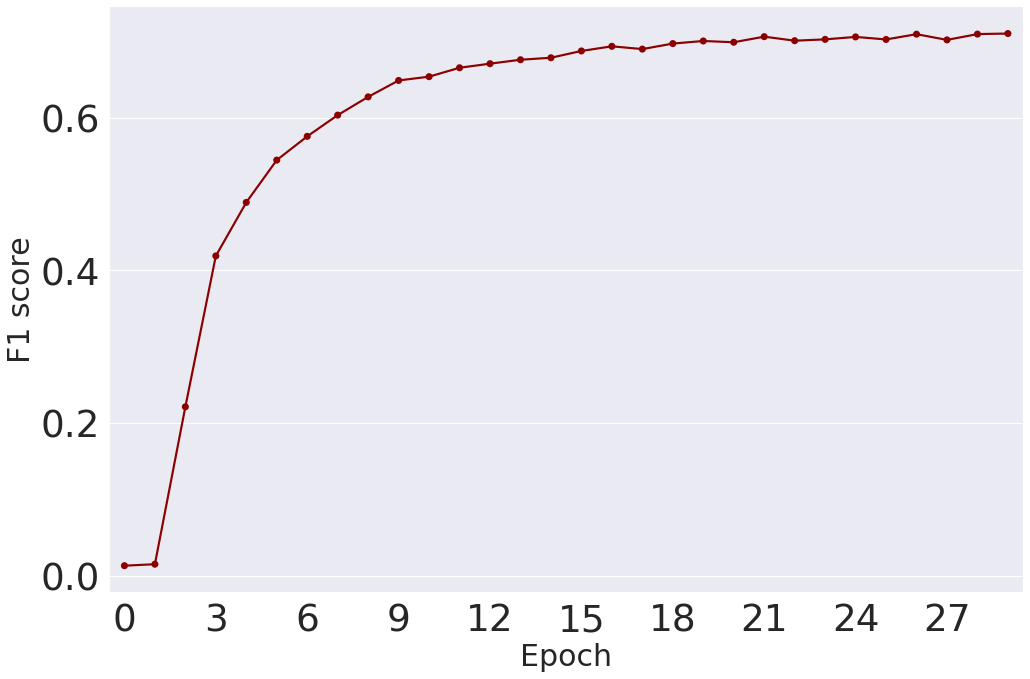} \\
    (c)
  \end{minipage}
  \caption{Train and validation loss: (a): pre-training (masked language modeling). (b): fine-tuning (class prediction). (c): F1 score for the validation dataset, during the fine-tuning step.}
  \label{fig:evals}
\end{figure}

To train the model, we used a single GPU. Table \ref{tbl:hp} shows the hyper-parameters for our model. We firstly pre-trained our model based on masked language modelling for 100 epochs \textit{(unsupervised)}. The pre-training step allows the model to discover common patterns in the SMILES strings by attempting to predict the masked tokens using the unmasked tokens. As discussed in Section \ref{sec:methodology}, the pre-trained model provides a proper starting point for training a model on a related desired task. This starting point incorporates the trained weights of the model. Furthermore, we validated the model on a separate dataset after each training epoch. The validation during training has no effect on model's trained weights, nevertheless, it helps in adjusting the model's hyper-parameters. Figure ~\ref{fig:evals} (a) illustrates the loss values for the train and validation sets during the pre-training phase. For the final multi-label classification task, we loaded the pre-trained model and trained it for 30 epochs with the class labels \textit{(supervised)}. Figure~\ref{fig:evals} (b) shows the train and validation loss during the fine-tuning step. Similarly, Fig.~\ref{fig:evals} (c) shows the F1 score for the validation dataset during the fine-tuning.

\section{Results and Evaluation}
\label{sec:eval}
\begin{figure}
  \centering
  \begin{minipage}[b]{0.3\textwidth}
    \centering
     \includegraphics[scale=0.1]{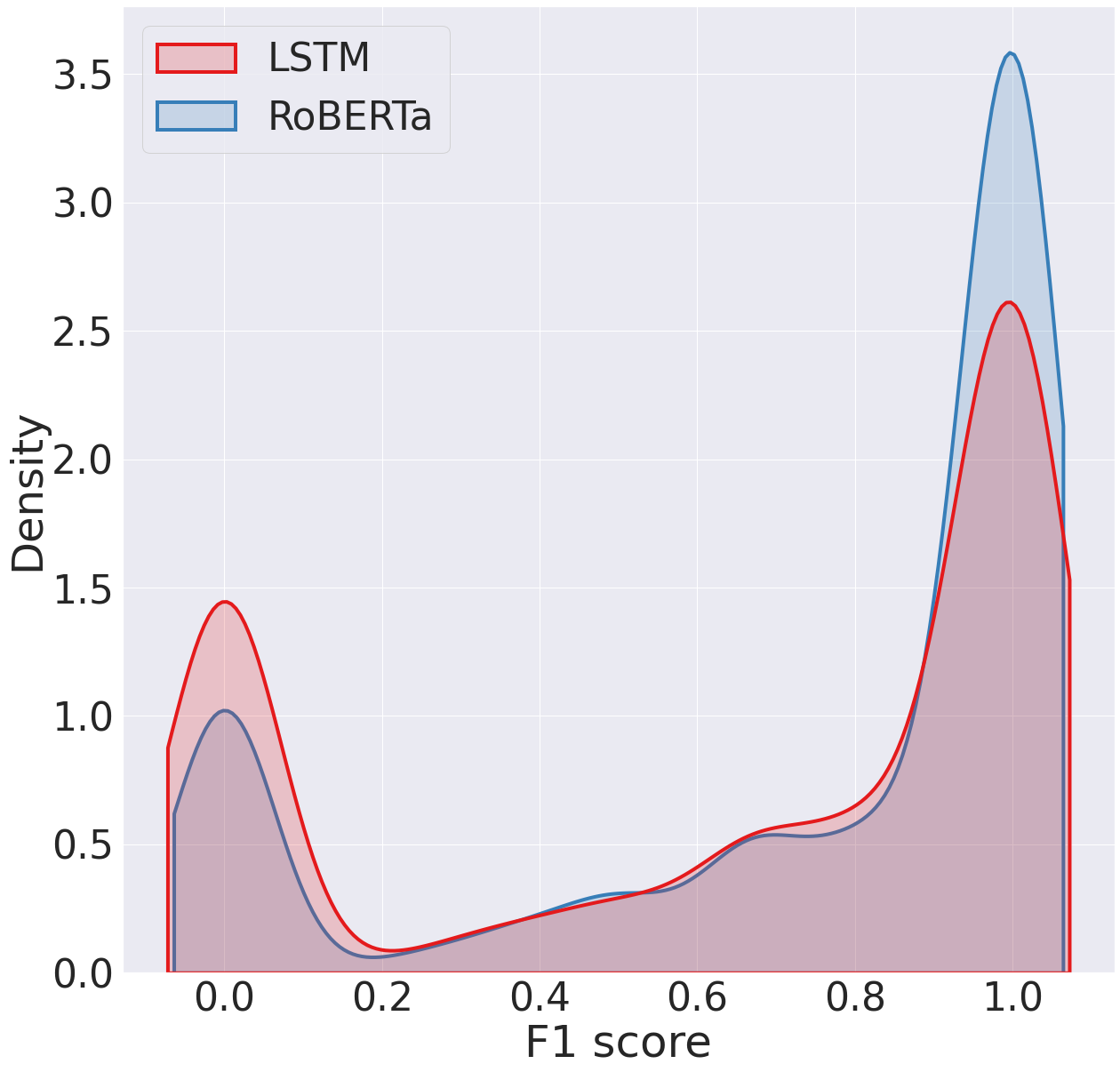} \\
    (a)
  \end{minipage}
  \hfill
  \begin{minipage}[b]{0.3\textwidth}
    \centering
    \includegraphics[scale=0.1]{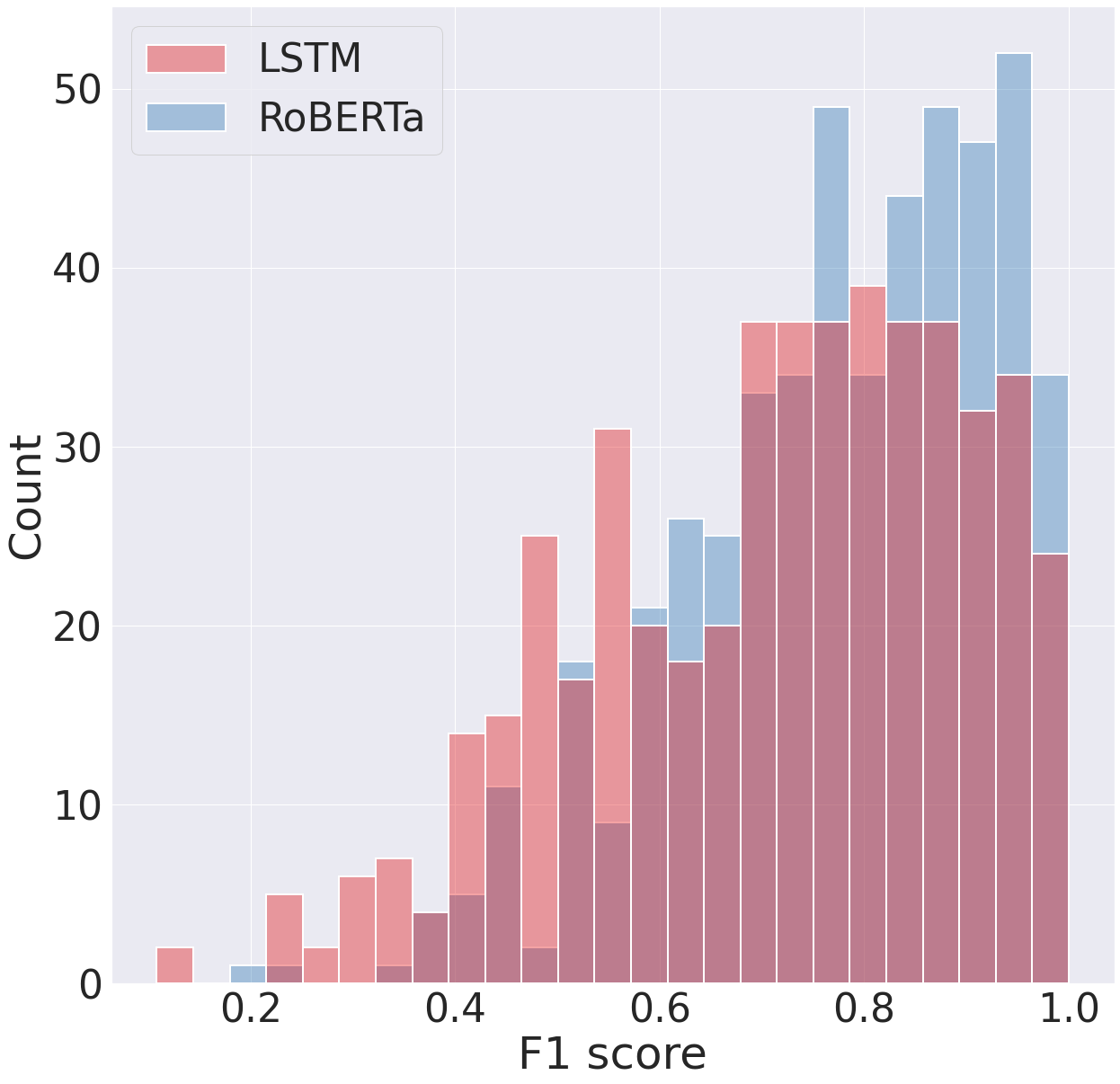} \\
    (b)
  \end{minipage}
  \hfill
  \begin{minipage}[b]{0.3\textwidth}
    \centering
    \includegraphics[scale=0.1]{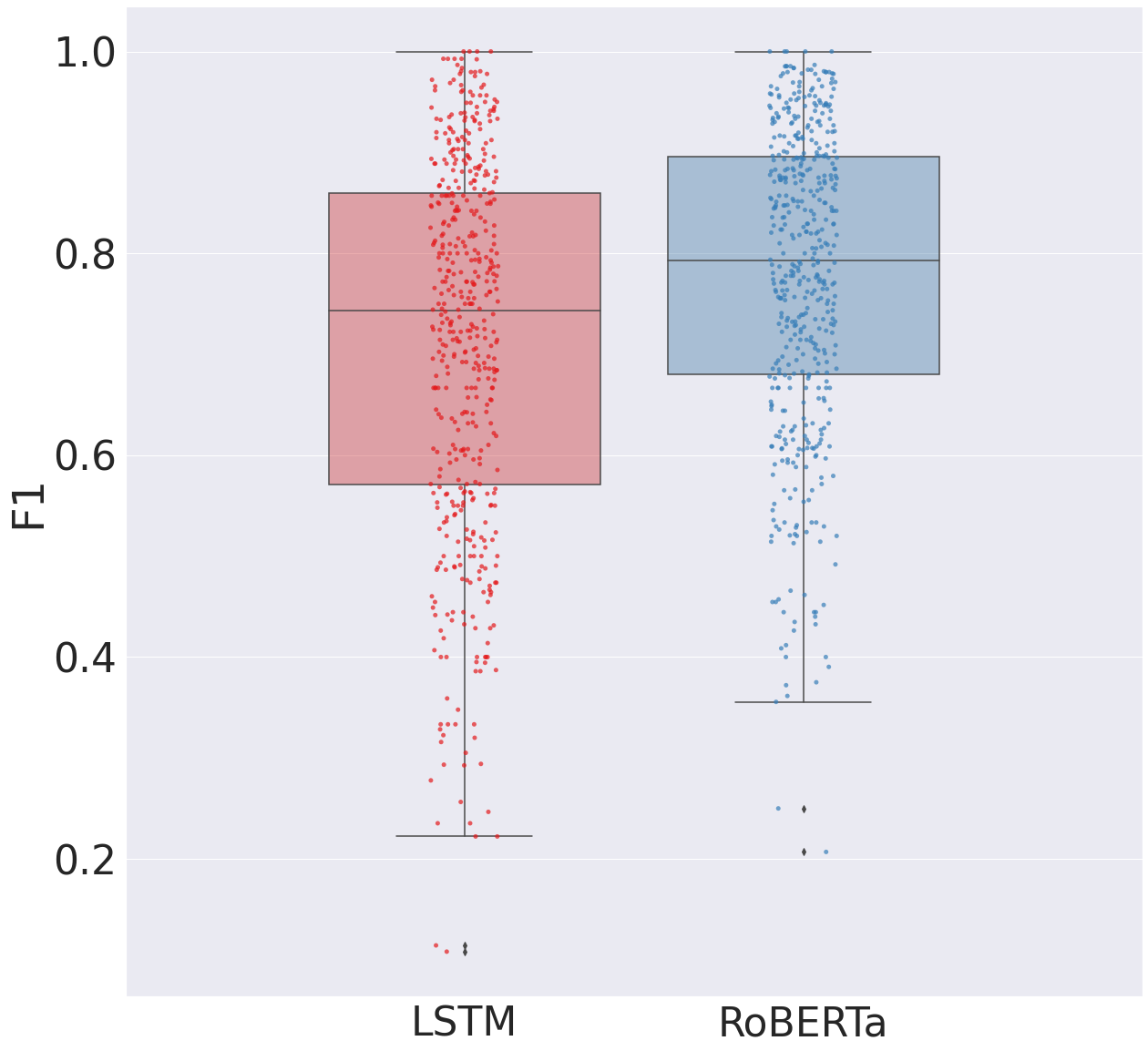} \\
    (c)
  \end{minipage}
  \caption{F1 score on test dataset. (a): Kernel density diagram based on the molecules. (b): Histogram diagram based on the classes. (c): Boxplots for the F1 scores of all 500 classes. A statistical test comparing the two class-wise F1 scores distributions yields a p-value of less than 0.001, indicating the distributions significantly differ.}
  \label{Fig:overal-f1}
\end{figure}

For our evaluations during and after training, we used the F1-score as the main measure. The F1 score may be computed in different ways depending on the averaging scheme: (1) \textit{samples}: calculates the F1 score for each molecule in the test dataset and then computes their average. (2) \textit{micro}: collects the total number of true positives, false positives, and false negatives and calculates the overall F1 score. (3) \textit{macro}: calculates the F1 score for each class and then computes their average. (4) \textit{weighted}: this averaging scheme is similar to the macro F1 score, but it calculates a weight for each class based on the number of true members in each class. Table \ref{tbl:scores} and Fig.~\ref{Fig:overal-f1} compare the results of the current model with the previously obtained results for the LSTM model from \cite{hastings2021learning}. We saw an improvement in performance both when we look at the distributions of values for the molecule-wise F1 scores (Figure~\ref{Fig:overal-f1}a) and for the class-wise F1 scores (Figure~\ref{Fig:overal-f1}b). A statistical comparison of the overall F1 score distributions shows that the difference in F1 scores is statistically significant ($p<0.001$, Figure~\ref{Fig:overal-f1}c). 

The raw output values of our model are the probabilities of a sigmoid function. Therefore, a threshold value must be applied to these probabilities to produce a binary vector, indicating the final classifications. These results are based on the classification threshold value of 0.5. The precision -- in our classification task -- shows the ability of the model to not wrongly assign a label to a molecule, while the recall score reflects the model's capability to discover all labels that were assigned to a molecule.

\begin{table}
\centering
\begin{center}
\begin{tabular}{lcccccccc}
\toprule
& \multicolumn{2}{c}{\textbf{Samples}}  &  \multicolumn{2}{c}{\textbf{Macro}} & \multicolumn{2}{c}{\textbf{Micro}}  &  \multicolumn{2}{c}{\textbf{Weighted}}  \\
\midrule
& \textbf{LSTM} & \textbf{RoBERTa} & \textbf{LSTM} & \textbf{RoBERTa} & \textbf{LSTM} & \textbf{RoBERTa} & \textbf{LSTM} & \textbf{RoBERTa} \\
\midrule
\textbf{F1} & 0.66 & 0.76 & 0.71 & 0.77 & 0.74 & 0.80 & 0.73 &  0.79 \\
\midrule
\textbf{Recall} & 0.66 & 0.75 & 0.68 & 0.76 & 0.70 & 0.78 & 0.70 & 0.78\\
\midrule
\textbf{Precision} & 0.67 & 0.77 & 0.77 & 0.80 & 0.79 & 0.82 & 0.79 & 0.82\\
\midrule
\textbf{ROC-AUC} & 0.83 & 0.87 & 0.84 & 0.89 & 0.85 & 0.88 & 0.85 & 0.89 \\
\bottomrule
\end{tabular}
\caption{The comparison of the scores achieved by two models.}
\label{tbl:scores}
\scriptsize
\end{center}
\end{table}

\begin{figure}
    \centering
    \includegraphics[scale=0.52]{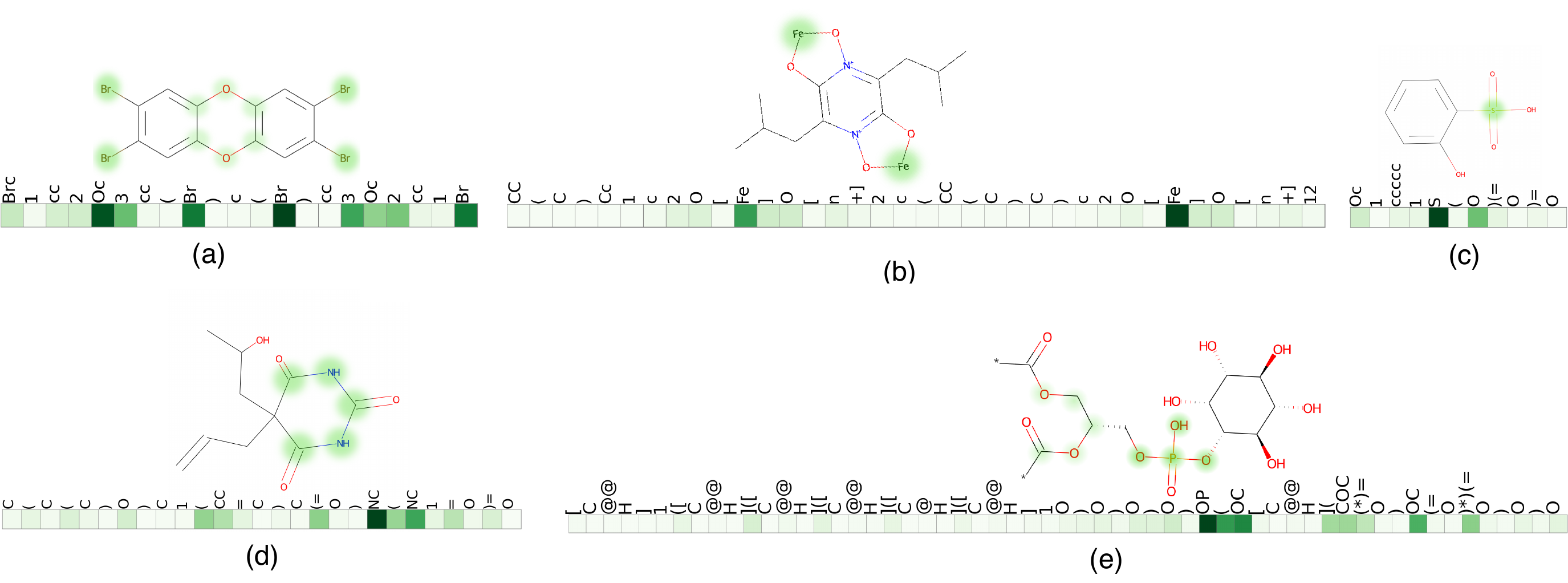}
    \caption{The model predicted class labels for these molecules by attending to influential sub-structures (highlighted in green): (a) organobromine compound (b) iron molecular entity (c) arenesulfonic acid (d) barbiturates (e) phosphatidylinositol}
    \label{Fig:mols_att}
\end{figure}

Self-attention in Transformer-based models enables the model to explore several locations in the input sequence to produce a better embedding for the tokens. As a result, the embeddings encode different contextual information for the same token in different positions (and different sequences). The architecture of the RoBERTa model contains a stack of Transformers' encoders, each consisting of multiple attention heads. Since the attention heads do not share parameters, each head learns a unique set of attention weights. Intuitively, attention weights determine the importance of each token for the embeddings of the next layers \cite{vig2006bertology}. In this sense, visualizing the attention weights of Transformer-based models helps to interpret the model with respect to the relative importance of different input items for making classifications \cite{vig2019multiscale}. While the benefit of attention visualization may be limited in explaining particular predictions, depending on the task, attention can be quite useful in explaining the model's overall predictions \cite{moradi2019interrogating, pruthi2019learning, serrano2019attention}. In fact, attention heads can reveal a wide variety of model behaviors and some of these heads may be more significant for model interpretation than others \cite{vig2019multiscale}. We examined how attention corresponds to different chemical structural elements, at both the token and molecule level. Figure~\ref{Fig:mols_att} shows the averaged attention weights of all heads in the last encoder of the model. The most attended sub-structures for each molecule are highlighted with green circles in the molecular graphs. It can be observed that often, most attention (darker green) is given to the heaviest atoms, for example \textit{bromine}, \textit{iron} and \textit{sulfur} in Fig.~\ref{Fig:mols_att} (a), (b) and (c) respectively. This corresponds to the broad principles of classification in organic chemistry as captured in ChEBI. The predicted classes in Fig.~\ref{Fig:mols_att} demonstrate that the model learned to assign appropriate labels to the chemical compounds. As illustrated in Fig.~\ref{Fig:mols_att} (d), the model assigned the \textit{barbiturates} class to the corresponding molecule, which class refers to the family of chemicals that contain a six-membered \textit{ring} structure, which was also the structural element given the most attention. Similarly, Fig.~\ref{Fig:mols_att} (e) shows that the model focused most on the \textit{phosphate} substructure when assigning the \textit{phosphatidylinositol} class to the molecule. 

\begin{figure}
    \centering
    \includegraphics[scale=0.53]{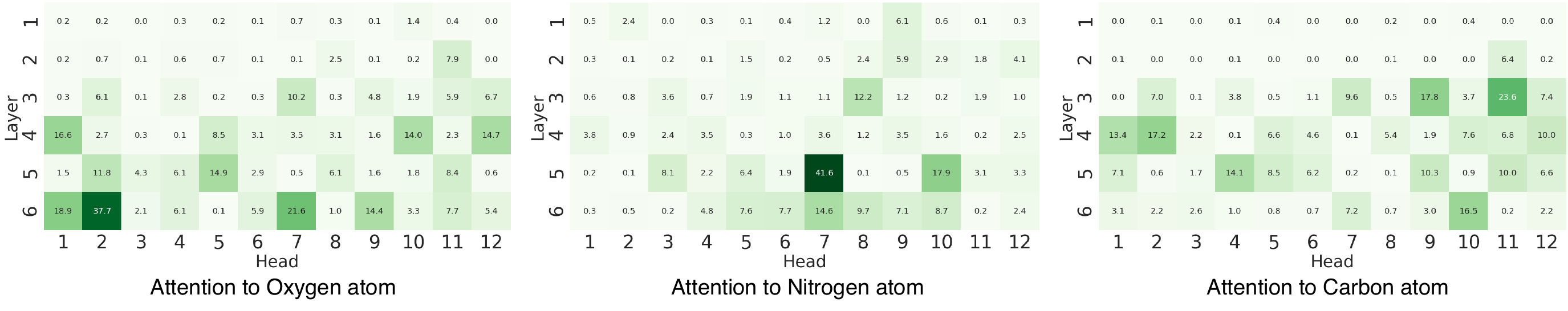}
    \caption{Each cell represents the percentage of all attentions (by each head) that was given to the corresponding token. For example, head 5-6 in (b) dedicated 41.6\% of its attention to the Nitrogen atom. }
    \label{Fig:overal_att}
\end{figure}
\begin{figure}
    \centering
    \includegraphics[width=\textwidth]{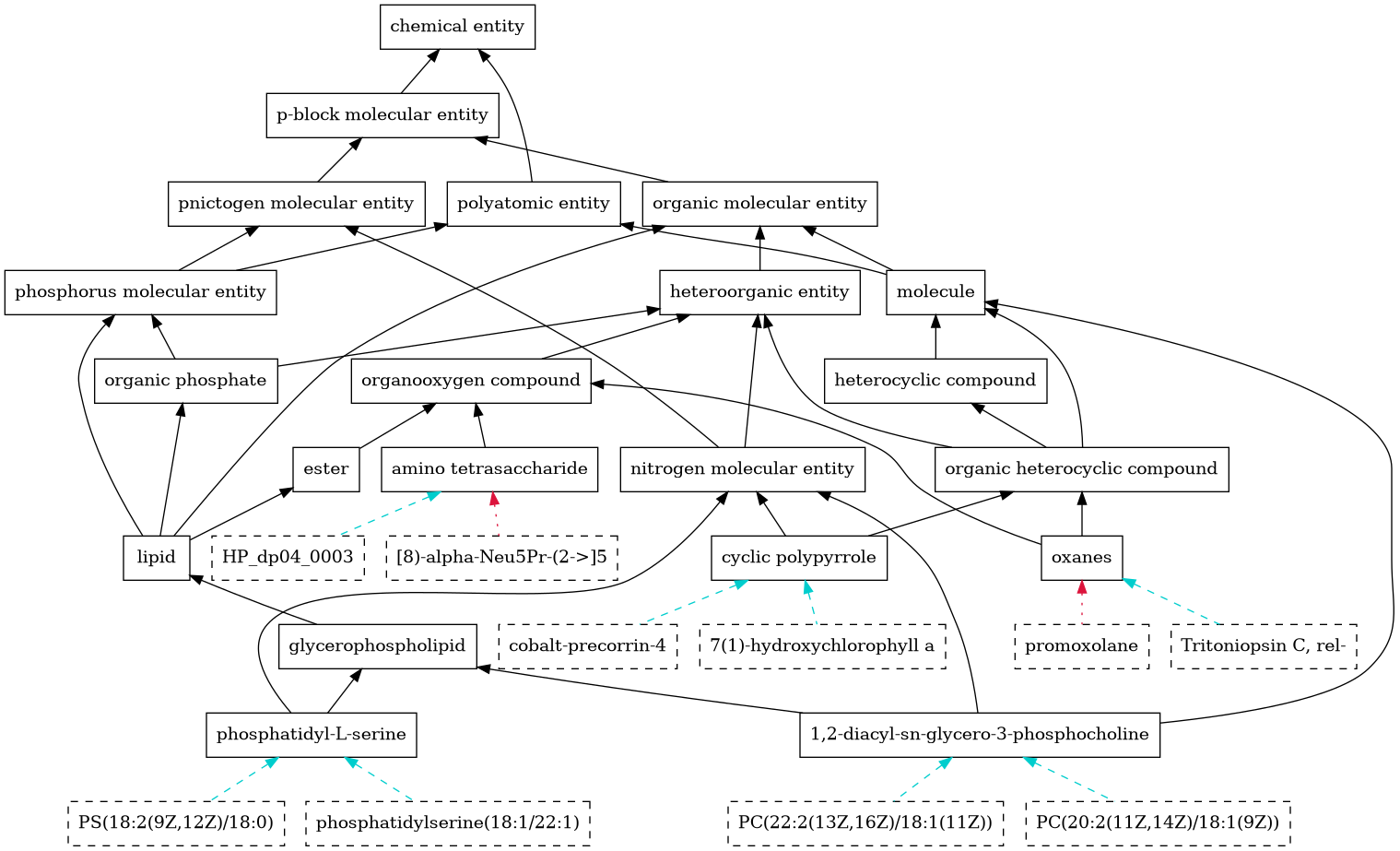}
    \caption{The extended ontology. Existing subsumption relations (black) have been enriched with new subclasses, shown with dashed borders. Correct subclass predictions are depicted with cyan, dashed arrows, while red, dotted arrows indicate misclassifications.}
    \label{Fig:ont_ext}
\end{figure}

The presented model takes a given class of molecules, represented by a SMILES string, and assigns the corresponding superclasses from the CHEBI ontology. ChEBI already makes use of an automated tool to extend its coverage beyond the manually curated core, namely ClassyFire. The model can be integrated into the ChEBI development process in the same way. The resulting system can then be used to integrate the given class into the ontology and translate the classification results into subsumption relations. Figure \ref{Fig:ont_ext} shows the result of this process. The implemented workflow, depicted in \ref{fig:ontolearn-arch}, is based on a model architecture and accompanying hyperparameters that have previously been used for chemical property prediction \cite{chithrananda2020chemberta}. The resulting model is then trained on the ontology data, which allows for the fully automated extension of the ChEBI ontology. 

\section{Discussion}
\label{sec:discussion}

ChEBI uses ClassyFire, a rules-based system, to extend its manually curated reference ontology to chemicals that are not yet covered. This approach has limitations, notably that
ClassyFire is structured around a different chemical ontology with only a partial mapping to ChEBI, and ClassyFire's rules are manually maintained. 
The deep-learning-based approach that we presented can overcome the limitations of rules-based approaches by allowing dynamic creation of classifiers based on a given existing ontology structure. Yet, for optimal applicability, the approach must meet certain quality criteria. Ozaki \cite{ozaki2020learning} defined six goals for ontology extension, which we use to structure our discussion of our results.

\noindent
\textbf{Handling of inconsistencies and noise} 
Our model is trained on information that originated solely from the ontology itself. This design decision eliminates external sources of inconsistencies and noise. 
The comparison of the F1 scores in the table in Fig.~\ref{Fig:overal-f1} shows that this classification outperforms the current state-of-the-art approaches - including the formerly leading LSTM-based model. In particular, for those chemical classes that were the most challenging in the previous approach, the current approach performed almost twice as well, as illustrated in Figure \ref{Fig:overal-f1} (b). 
It should be noted that there nevertheless remain some chemical classes that perform worse than others. For example, classes that are based on cyclic structures pose challenges, as their information may be scattered around the respective SMILES strings. Alternative input formats and network architectures may be explored in the future to better handle these structures. 
The model may also benefit from a larger amount of data. The distribution of class memberships depicted in Figure \ref{Fig:dataset} indicates that the dataset features some classes far more often than others. These classes are more prominent, often by virtue of being higher in the ontology subclass hierarchy and, therefore, represent broader classes of chemicals that may share members with other classes. Such an imbalance can skew the training in favour of those classes. Different sampling and regularization techniques may be explored in the future to address this issue.

\noindent
\textbf{Unsupervised learning} The presented approach is a variant of ontology extension. The ontology is therefore a mandatory input, from which the information that is needed for the ontology extension is extracted. The resulting dataset does include labels for each molecule. Strictly speaking, it is thus a supervised learning approach. However, these labels are extracted fully automatically from the input -- the ontology. Therefore, no additional annotation by experts or other manual data pre-processing is necessary.

\noindent
\textbf{Human interaction} As the ontology is extended automatically, no  interaction is required. 

\noindent
\textbf{Expressivity} The system extends the given ontology using the same ontology language that has been used to build it. ChEBI is developed as an OWL ontology, which comes with expressive OWL-DL semantics.

\noindent
\textbf{Interpretability} The formerly best classifier was based on an LSTM architecture. This approach outperformed ClassyFire, but this performance came with a disadvantage: The reason for a specific classification was not transparent. This is problematic, because the experts that check the ontology extension need insights into the system's decision processes in order to evaluate the classifications. An explainable approach is therefore crucial. The attention mechanism of the RoBERTa architecture that has been used in the present approach helps to address this issue. Attention weights can be seen as a measure of how much focus is put on an individual token. A homogenous distribution of attention shows that nothing has been focused in particular, whilst high attention on a head shows that a particular token had a high impact. Figure \ref{Fig:overal_att} shows that carbon atoms, which are very common in organic chemistry, trigger a low general focus. At the same time, a high focus is put on oxygen atoms, that often indicate functional groups of high classificatory relevance, such as carboxy groups.
Figure \ref{Fig:mols_att} shows which parts of a particular molecule have been focused on during the classification process. This information can be used to explain the decisions made by the model, raise trust in the prediction system, and aid the experts during the ontology extension process.

\noindent
\textbf{Efficiency} In  \cite{ozaki2020learning} `efficiency' is defined as the time it takes  to build the ontology. Once the model is fully trained, the  classification which leads to the ontology extension only takes a few minutes. As an example, classification of 6,569 chemical entities in our test dataset took around 10 minutes. While extending the ontology itself is fast, the training of the model requires more time.  Training is divided in pre-training and fine-tuning. The pre-training with 100 epochs took around 10 hours. This time is only invested once, and thereafter a pre-trained model can be fine-tuned repeatedly for several large sets of molecules and their corresponding classes comparatively quickly. Our final fine-tuning for 30 epochs took around 2 hours.

\smallskip
This analysis shows that the presented approach achieves the goals of ontology learning stipulated in \cite{ozaki2020learning}. One additional issue that needs to be addressed is \textit{applicability}. At the heart of the presented approach is a neural network that is trained based on the annotations of the ontology. In the same way as any text analysis approach to ontology generation is dependent on the existence of suitable text corpora, our approach requires that the ontology contains enough information to train a model to predict the superclasses of a new class. ChEBI is an ideal use case, because SMILES annotations provide rich, structured information that we could harness for training the model. Another potential application domain for our approach in biology are proteins, which are also classified based on structures, features of which can be annotated in the relevant ontology. 
Moreover, our approach is not limited to ontologies with structural information represented in annotations. E.g., for ontologies in material science one could consider training the model based on the physical properties (e.g., density, hardness, thermal conductivity), which are typically represented using data properties.  
In short, our approach to ontology extension is applicable to reference ontologies that associate classes with sufficient information that a neural network may learn the classification criteria that the ontology developers are using.   

\section{Conclusion and Future Work}

We have presented a novel approach to the problem of ontology extension, applied to the chemical domain. Instead of extending the ontology using external resources, we created a model using the ontology's own structured annotations. This Transformer-based model can not only classify previously unseen chemical entities (such as molecules) into the appropriate classes, but also provides information about relevant aspects of its internal structure on which the decision is based. At the same time, it was able to outperform previously existing approaches to ontology-based chemical classification in terms of predictive performance. 

However, the trained model still struggles with several chemical classes that depend on specific structural features. E.g, classes that exhibit cyclic structures are often found in the lower quantile of classification quality. This behaviour can be traced back to the way molecules are encoded into the SMILES notation. This weakness might be addressed by using  architectures  that operate directly on the molecular structures, such as Graph Neural Networks \cite{scarselli2008graph}.

We have illustrated our approach by applying it to the chemical domain, but as we discussed in Section~\ref{sec:discussion}, the approach is applicable to any ontology that contains classes that are annotated with information that is relevant to their position in the class hierarchy. 

While our approach supports an automatic extension of an ontology, it can also be used in a semi-automated fashion to help  ontology developers in their manual curation of the ontology.
Since the model is trained based on the content of a manually curated ontology, improving and extending this ontology will lead to better quality training data and, thus, enable better predictions. Hence, there is a potential for a positive feedback loop between manual development and  the AI-based extension. 

One limitation of our current approach is that it 
does not use most of the logical axioms of the ontology 
during the learning process. 
The logical axioms within the ontology could be used to detect possible inconsistencies between the predicted classes and the ontology's axioms. A logic-based framework could then be employed to detect those results that were most likely mis-classifications.
Another strategy to address this gap would be to represent the axioms in the form of Logical Neural Networks \cite{riegel2020logical} in order to detect possible inconsistencies already in the learning process and to penalise them accordingly. Overall, there is still a pressing need for research in the field of (semi-)automatic ontology extension. Here, the growing field of neuro-symbolic integration can serve as the interface between formal ontologies and the power of deep learning. The possibility of incorporating explanations may further the understanding of the inner workings of artificial intelligence systems and, therefore, raise trust in these systems.

\bibliography{bibliography}

\appendix

\end{document}